%% file: main.tex
\documentclass{article}
\usepackage{amssymb}
\usepackage[dvipsnames,table]{xcolor}
\usepackage{amsmath}
\usepackage[preprint]{corl_2025} 

\usepackage[nameinlink,capitalise]{cleveref} 
\crefname{lstlisting}{listing}{listings}
\Crefname{lstlisting}{Listing}{Listings}
\usepackage{multirow}
\usepackage{etoc}
\usepackage{listings}
\usepackage[font=footnotesize]{caption}
\usepackage[font=footnotesize]{subcaption}
\usepackage{wrapfig}

\definecolor{beat}{HTML}{e3e3e3}
\newcommand{\hl}[1]{%
  \begingroup
    \setlength{\fboxsep}{0pt}%
    \colorbox{beat}{\strut#1}%
  \endgroup
}
\lstdefinestyle{promptstyle}{
    backgroundcolor=\color{gray!8},
    basicstyle=\scriptsize\ttfamily\color{black!90},
    breakautoindent=false,
    breakindent=0pt, 
    breakatwhitespace=true,
    breaklines=true,
    captionpos=b,
    keepspaces=true,
    showspaces=false,
    showstringspaces=false,
    showtabs=false,
    frame=single,
    rulecolor=\color{gray!40},
    framesep=3pt,
    frameround=tttt,
    framexleftmargin=6pt,
    xleftmargin=8pt,
    xrightmargin=8pt,
    tabsize=2,
    linewidth=0.98\textwidth,
    fontadjust=true,
    numbers=none,
    aboveskip=0.8\baselineskip,
    belowskip=0.8\baselineskip,
    columns=flexible,
    upquote=true,
    inputencoding=utf8,
    extendedchars=true,
    lineskip=-0.1pt,
    resetmargins=true,
    moredelim=[s][\color{blue!70!black}]{\{}{\}},
}

\lstdefinestyle{heuristicstyle}{
    backgroundcolor=\color{gray!5},
    commentstyle=\color{green!60!black},
    keywordstyle=\color{blue!70!black}\bfseries,
    numberstyle=\tiny\color{gray!70},
    stringstyle=\color{purple!70!black},
    basicstyle=\scriptsize\ttfamily\color{black},
    breakatwhitespace=false,
    breaklines=true,
    captionpos=b,
    keepspaces=true,
    showspaces=false,
    showstringspaces=false,
    showtabs=false,
    tabsize=4,
    frame=single,
    rulecolor=\color{gray!50},
    framesep=3pt,
    frameround=tttt,
    numbersep=6pt,
    xleftmargin=10pt,
    xrightmargin=10pt,
    aboveskip=1.0\baselineskip,
    belowskip=1.0\baselineskip,
    upquote=true,
    columns=flexible,
    keepspaces=true,
    mathescape=true,
    escapeinside={(*@}{@*)},
    morecomment=[l]\#,
    morekeywords={import, from, as, def, class, return, yield, for, while, if, elif, else, try, except, finally, with, lambda, pass, break, continue, and, or, not, is, in, raise, assert},
    emph={self, None, True, False, np, pd, plt, torch, tf, sklearn},
    emphstyle=\color{orange!80!black}\bfseries,
    literate=
        {-}{-}1
        {=>}{$\Rightarrow$ }3
        {->}{$\rightarrow$ }3
        {...}{$\ldots$ }3,
    inputencoding=utf8,
    extendedchars=true,
    lineskip=-0.1pt,
    fontadjust=true
}

\usepackage{graphicx}
\usepackage{booktabs}
\usepackage{multirow}
\usepackage{array}
\newcommand{\our}{{\textsc{TrajEvo}}}

\usepackage{booktabs} 

\title{\our{}: Designing Trajectory Prediction Heuristics via LLM-driven Evolution}

\author{
  Zhikai Zhao\thanks{Joint first authors.}\hspace{1.3mm}$^{,1}$ \hspace{2mm} Chuanbo Hua$^{*,1,2}$ \hspace{2mm} Federico Berto$^{*,1,2}$ \vspace{1mm} \\ \textbf{Kanghoon Lee}$^{1}$ \hspace{2mm} \textbf{Zihan Ma}$^{1}$ \hspace{2mm}  \textbf{Jiachen Li}$^{3}$  \hspace{2mm} \textbf{Jinkyoo Park}$^{1,2}$ \vspace{2mm} \\ 
  $^{1}$KAIST \hspace{1mm} $^{2}$OMELET \hspace{1mm} $^{3}$ UC Riverside \hspace{1mm} AI4CO\thanks{Work made with contributions from the AI4CO open research community.} \\
}

\begin{document}
\maketitle


\begin{abstract}
%
    %
Trajectory prediction is a crucial task in modeling human behavior, especially in fields as social robotics and autonomous vehicle navigation. Traditional heuristics based on handcrafted rules often lack accuracy, while recently proposed deep learning approaches suffer from computational cost, lack of explainability, and generalization issues that limit their practical adoption. In this paper, we introduce \our{}, a framework that leverages Large Language Models (LLMs) to automatically design trajectory prediction heuristics. \our{} employs an evolutionary algorithm to generate and refine prediction heuristics from past trajectory data. We introduce a Cross-Generation Elite Sampling to promote population diversity and a Statistics Feedback Loop allowing the LLM to analyze alternative predictions. Our evaluations show \our{} outperforms previous heuristic methods on the ETH-UCY datasets, and remarkably outperforms both heuristics and deep learning methods when generalizing to the unseen SDD dataset. \our{} represents a first step toward automated design of fast, explainable, and generalizable trajectory prediction heuristics. We make our source code publicly available to foster future research at \url{https://github.com/ai4co/trajevo}.

\end{abstract}



\section{Introduction}
\label{sec:introduction}

Trajectory prediction is an important cornerstone of intelligent autonomous systems \citep{madjid2025trajectory}, with several real-world applications, including autonomous driving \citep{wang2024escirl}, safety in industry \citep{8107677}, robotics navigation \citep{vishwakarma2024adoption}, and planning \citep{li2022human}. The inherent stochasticity of these settings necessitates systems capable of accurately detecting and processing data with both temporal and spatial precision \citep{nakamura2024not}. For example, in the navigation of self-driving vehicles, the ability to accurately predict the future trajectories of pedestrians is essential for ensuring safety in the complex urban environment without collision with pedestrians; industry and indoor operational robots need precise predictions to collaborate safely with humans in shared workspaces, avoiding harming the people around \citep{mavrogiannis2023core,mahdi2022survey}.

However, accurately predicting the movement patterns of multiple agents like pedestrians or vehicles is fundamentally challenging. Human motion is complex, often involving nonlinear behaviors, rapid directional changes, and spontaneous decisions influenced by individual factors like habits or urgency. Furthermore, agents interact dynamically, adjusting paths to avoid collisions, responding to traffic, or moving cohesively in groups, leading to a high degree of stochasticity \citep{amirian2020opentraj}. Early attempts to model these behaviors relied on heuristic methods, such as the Social Force model \citep{helbing1995social}, which represents motivations and interactions as physics-inspired forces \citep{helbing1995social,5509779,zanlungo2011social,farina2017walking,chen2018social}, constraint-based approaches defining collision-free velocities \citep{4543489, Ma2018EfficientRC}, or agent-based simulations of decision-making \citep{5995468}. While interpretable, these handcrafted heuristics often struggle with accuracy. Their reliance on handcrafted rules and parameters makes them difficult to design and tune effectively for complex, dynamic scenarios, frequently resulting in limited accuracy and poor generalization.

\begin{figure}[t]
    \centering
\includegraphics[width=0.75\textwidth]{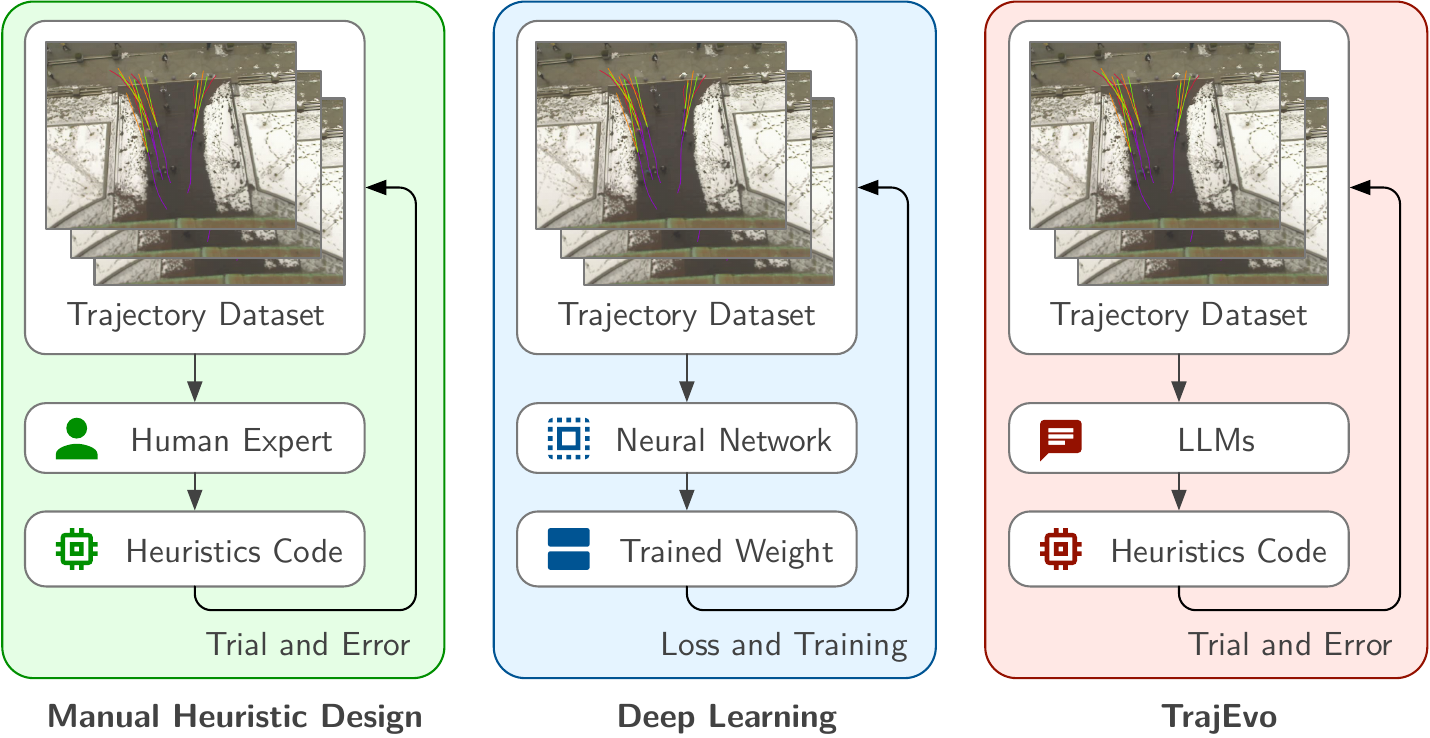}
    \caption{Motivation for \our{}. Traditional manual heuristic design (left) is based on human experts with trial and error. Deep learning (center) generates better predictions but requires significant computational resources, generates black-box models, and struggles with generalization. \our{} (right) automates the design process of heuristics via evolutionary algorithms, generating novel trajectory prediction heuristics.}
    \label{fig:motivation}
\end{figure}

Deep learning methods have been introduced as a powerful alternative to heuristics \citep{alahi2016social,salzmann2020trajectron++,li2022evolvehypergraph}. Social-LSTM \citep{alahi2016social} stands out as an early, influential model leveraging LSTMs for this task. Following this, a wide variety of new learning approaches, including GNN \citep{rainbow2021semantics,rainbow2021semantics}, and GANs \citep{gupta2018social,dendorfer2021mg} have been explored to further improve prediction accuracy. Transformer-based frameworks \citep{kim2025guide,bae2024can} and diffusion models \citep{yang2024uncovering,gu2022stochastic,fu2025moflow}  have recently emerged as a promising paradigm for capturing complex dependencies in human movement patterns. Such methods, however, suffer from several issues in practice, including the need for powerful hardware that restricts real-time use cases \citep{itkina2023interpretable,jiang2025survey}, overall lack of interpretability of their predictions \citep{liu2024traj,cai2024pwto}, and robustness in out-of-distribution settings \citep{korbmacher2022review, rudenko2020human}. For this reason, many heuristics are practically used instead in safety-critical applications demanding consistent real-time performance \citep{phong2023truly}.

In this paper, we pose the following research question -- \emph{Can we automatically design accurate trajectory prediction heuristics to bridge the gap with deep learning?}

Inspired by recent research exploring combinations of Large Language Models (LLMs) with Evolutionary Algorithms (EAs) for automated algorithm design \citep{liu2023algorithm,dat2025hsevo,ye2024reevo,chen2024uber,liu2024evolution,yuksekgonul2025optimizing,liu2024llm4ad,liu2024systematic,wu2024evolutionary,zhang2024understanding}, we propose a new approach to automatically generate effective trajectory prediction heuristics. We hypothesize that coupling the generative and reasoning capabilities of LLMs with the structured search of EAs can overcome the limitations of manual heuristic design and bridge the gap with deep learning models.

In this paper, we introduce \our{} (\textbf{Traj}ectory \textbf{Evo}lution), a new framework designed to achieve this goal. \our{} leverages LLMs within an evolutionary loop to iteratively generate, evaluate, and refine trajectory prediction heuristics directly from data. Our main contributions are as follows:
\begin{itemize}
    \item We present \our{}, the first framework, to the best of our knowledge, that integrates LLMs with evolutionary algorithms specifically for the automated discovery and design of fast, explainable, and robust trajectory prediction heuristics.
    \item We introduce a Cross-Generation Elite Sampling strategy to maintain population diversity and a Statistics Feedback Loop that enables the LLM to analyze heuristic performance and guide the generation of improved candidates based on past trajectory data.
    \item We demonstrate that \our{} generates heuristics significantly outperforming prior heuristic methods on the standard ETH-UCY benchmarks, and exhibit remarkable generalization, surpassing both traditional heuristics and state-of-the-art deep learning methods on the unseen SDD dataset, while remaining computationally fast and interpretable.
\end{itemize}

\section{Related Work}
\label{sec:related_work}

\paragraph{Heuristic Methods for Trajectory Prediction}
Traditional heuristic approaches provide interpretable frameworks for trajectory prediction but often with accuracy limitations. The Constant Velocity Model (CVM) and its sampling variant (CVM-S) \citep{scholler2020constant} represent foundational baselines that assume uniform motion patterns. More sophisticated approaches include Constant Acceleration \citep{polychronopoulos2007sensor}, CTRV (Constant Turn Rate and Velocity) \citep{lu2021ctrv}, and CSCRCTR \citep{s140305239}, which incorporate different kinematic assumptions. The Social Force Model \citep{helbing1995social} pioneered physics-inspired representations of pedestrian dynamics, with numerous extensions incorporating social behaviors \citep{zanlungo2011social, farina2017walking, chen2018social} and group dynamics \citep{helbing1997modeling}. While these heuristics offer computational efficiency and explainability, they typically struggle with complex real-world, multi-agent interactions due to their limited expressiveness and reliance on manually defined parameters -- limitations our proposed \our{} framework specifically addresses through automatic heuristic generation.

\paragraph{Learning-based Trajectory Prediction}
Deep learning has substantially advanced trajectory prediction accuracy at the cost of computational efficiency and interpretability. Social-LSTM \citep{alahi2016social} is a seminal work that introduced a social pooling mechanism to model inter-agent interactions, while Social-GAN \citep{gupta2018social} leveraged generative approaches to capture trajectory multimodality. Graph-based architectures like STGAT \citep{Huang_2019_ICCV} and Social-STGCNN \citep{mohamed2020social} explicitly model the dynamic social graph between pedestrians. Recent approaches include Trajectron++ \citep{salzmann2020trajectron++}, which integrates various contextual factors, MemoNet \citep{xu2022remember} with its memory mechanisms, EigenTrajectory \citep{bae2023eigentrajectory} focusing on principal motion patterns, and MoFlow \citep{fu2025moflow} employing normalizing flows for stochastic prediction. While these methods achieve high in-distribution accuracy, they frequently require significant computational resources, lack interpretability, and often struggle with out-of-distribution generalization -- three critical challenges that motivate our \our{} framework's design.

\paragraph{LLMs for Algorithmic Design}
Large Language Models have recently demonstrated remarkable capabilities in algorithmic design and code generation tasks. LLMs have been successfully applied to robotics policy generation \citep{liang2023code}, code synthesis \citep{chen2023natural}, and optimization \citep{xia2023natural}. Most relevant to our work are approaches that leverage LLMs within evolutionary frameworks to design algorithms and heuristics \citep{romera2024mathematical, liu2024evolution, ye2024reevo}. These frameworks harness LLMs' reasoning capabilities to generate and refine algorithmic solutions through iterative evolutionary processes. Our work, \our{}, extends these concepts to the domain of trajectory prediction as the first approach that combines LLMs with evolutionary algorithms to automatically discover computationally efficient, interpretable, and generalizable trajectory prediction heuristics to bridge the gap between traditional handcrafted approaches and sophisticated neural methods.

\section{\our{}}
\label{sec:methodology}

\begin{figure}[t]  
    \centering
\includegraphics[width=0.85\textwidth]{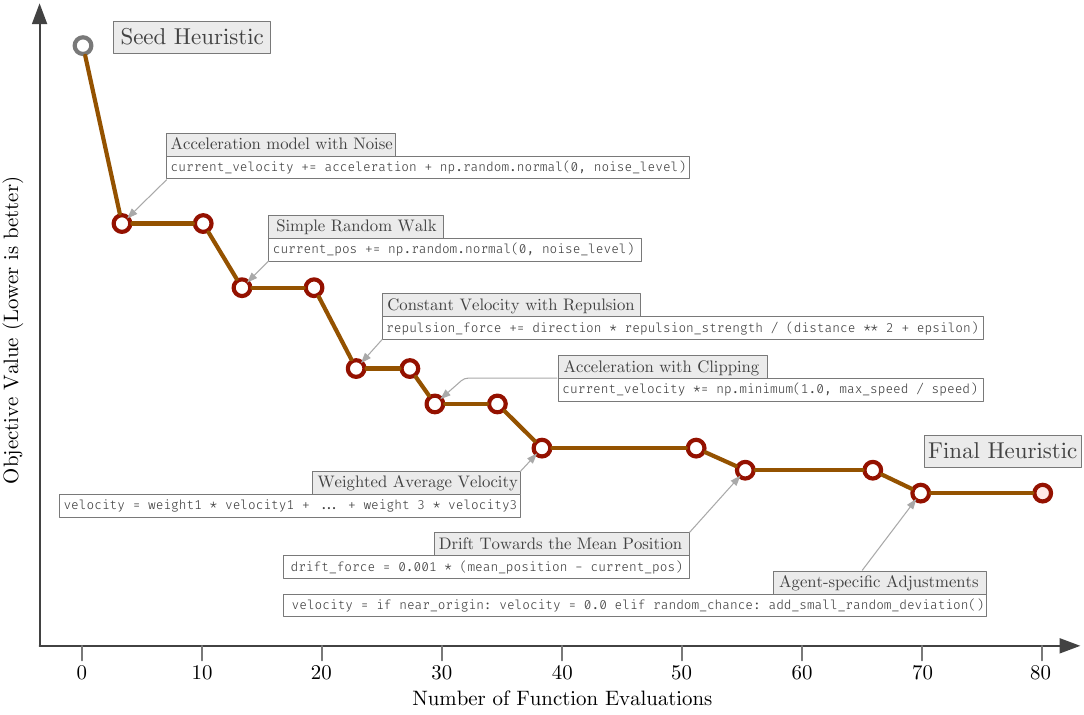}
    \caption{Example evolution of trajectory prediction heuristics with \our{}.}
    \vspace{-3mm}
    \label{fig:trajevo-catchy}
\end{figure}



\subsection{Problem Definition}
\label{subsec:problem-definition}

\paragraph{Task}
We address multi-agent trajectory prediction. The task is, for each agent $i$, to predict its future path $Y_i = (P_i^{T_{\text{obs}}+1}, \dots, P_i^{T_{\text{obs}}+T_{\text{pred}}})$ given its observed history $H_i = (P_i^1, \dots, P_i^{T_{\text{obs}}})$, where $P_i^t \in \mathbb{R}^2$ is the position at timestep $t$. Predictive heuristics are evolved (i.e., trained) on a training set $\mathcal{D}_{\text{train}}$ and evaluated on a test set $\mathcal{D}_{\text{test}}$.

\paragraph{Metrics}
We evaluate heuristics using the standard multi-sample protocol \citep{gupta2018social}. For each input scene, the heuristic generates $K=20$ distinct sets of future trajectories $\{\{\hat{Y}_i^k\}_{i=1\dots N}\}_{k=1\dots K}$, where each set $k$ contains predicted paths for all $N$ agents. To select the best prediction, we identify the single set index $k^*$ that yields the smallest overall error. We use as metrics the minimum Average Displacement Error ($\text{minADE}_{20}$) and minimum Final Displacement Error ($\text{minFDE}_{20}$) are then computed using the trajectories $\{\hat{Y}_i^{k^*}\}_{i=1 \dots N}$ corresponding to the single best-performing set index $k^*$. This multi-sample approach accounts for the inherent multimodality of human movement by assessing a prediction method's ability to generate at least one possible future scenario. 

\paragraph{Combined Objective}
We use a weighted sum of the above as the combined objective value $J$ to be minimized, defined as a weighted sum of the above metrics: $J = w_{\text{ADE}} \times \text{minADE}_{20} + w_{\text{FDE}} \times \text{minFDE}_{20}$ with $w_{\text{ADE}}$ and $w_{\text{FDE}}$ 0.6 and 0.4, respectively.

\subsection{Evolutionary Framework}
\label{subsec:evolutionary-framework} 

Our evolutionary framework adapts the Reflective Evolution approach \citep{ye2024reevo}. We leverage Large Language Models (LLMs) to implement core genetic algorithm operators -- initialization, crossover, and mutation -- which are guided by reflective processes analyzing heuristic performance.

\paragraph{Initial Population}
The process starts by seeding the generator LLM with a task specification (including problem details, input/output format, and the objective $J$) alongside a basic heuristic like the Constant Velocity Model (CVM) \citep{scholler2020constant}. The LLM then generates an initial population of $N$ diverse heuristics, providing the starting point for the evolutionary search.


\paragraph{Selection for Crossover}
Parents for crossover are chosen from successfully executed heuristics in the current population. The selection balances exploration and exploitation: 70\% of parents are selected uniformly at random from successful candidates, while 30\% are chosen from the elite performers (those with the lowest objective $J$).

\paragraph{Reflections}
\our{} employs two types of reflection as in \citet{ye2024reevo}. \textit{Short-term reflections} compare the performance of selected crossover parents, offering immediate feedback to guide the generation of offspring. \textit{Long-term reflections} accumulate insights across generations, identifying effective design patterns and principles to steer mutation and broader exploration. Both reflection mechanisms produce textual guidance (i.e. ``verbal gradients'' \citep{pryzant2023automatic}) for the generator LLM.

\paragraph{Crossover}
This operator creates new offspring by combining code from two parent heuristics. Guided by short-term reflections that compare the `better' and `worse' performing parent, the LLM is prompted to mix their effective ``genes", enabling the emergence of potentially superior heuristics.

\paragraph{Elitist Mutation} 
The mutation operator mutates the elitist (best found so far) heuristic. In \our{}, this involves the generator LLM modifying an elite heuristic selected. This mutation step is informed by the insights gathered through long-term reflections.

\subsection{Cross-Generation Elite Sampling}
\label{subsec:cross-generation-elite-sampling}

Evolving effective heuristics for complex tasks like trajectory prediction poses a significant search challenge, where standard evolutionary processes can easily get trapped in local optima \citep{osuna2018runtime}. Simple mutations often yield only incremental improvements, failing to explore sufficiently diverse or novel strategies. To address this limitation and enhance exploration, we introduce Cross-Generation Elite Sampling (CGES), a core component influencing the mutation step within the \our{} framework.

\input{figures/cross-generation-elite-sampling} 

In contrast to typical elitism focusing on the current generation's best, CGES maintains a history archive of high-performing heuristics accumulated across all past generations. Specifically, CGES modifies how the elite individual targeted for mutation is selected: instead of necessarily choosing the top performer from the current population, the heuristic designated to undergo mutation is sampled by CGES directly from this history. This sampling uses a Softmax distribution based on the recorded objectives $J$ of the historical elites, thus prioritizing individuals that have proven effective previously. By potentially reintroducing and modifying diverse, historically successful strategies during the mutation phase, CGES significantly improves the exploration capability, facilitating escape from local optima and the discovery of more robust heuristics, as demonstrated in \cref{fig:cross-generation-elite-sampling}.

\subsection{Statistics Feedback Loop}
\label{subsec:statistics-feedback-loop}

While the objective $J$ measures overall quality, it does not reveal \textit{which} of a heuristic's diverse internal prediction strategies are actually effective. To provide this crucial insight for effective refinement, \our{} incorporates a Statistics Feedback Loop.


\input{figures/trajectory-analysis} 

This loop specifically analyzes the contribution of the $K=20$ distinct trajectory prediction sets (indexed $k=0\dots19$) generated by a heuristic. After evaluation, we compute a key statistic: a distribution showing how frequently each prediction index $k$ delivered the minimum ADE for individual trajectory instances across the dataset (\cref{fig:trajectory-analysis}, left panel). This directly highlights the empirical utility of the different diversification strategies associated with each index $k$. This statistical distribution, together with the heuristic's code, is fed to the reflector and mutation LLMs to identify which strategies ($k$) contribute most effectively to performance (\cref{fig:trajectory-analysis}, right panel). Such feedback provides actionable guidance to the generator LLM, enabling specific improvements to the heuristic's multi-prediction generation logic based on the observed effectiveness of its constituent strategies. A qualitative example of the full \our{} evolution is provided in \cref{fig:trajevo-catchy}.


\section{Experiments}
\label{sec:result}

\subsection{Experimental Setup}

\paragraph{Datasets}
We evaluate \our{} on the ETH-UCY benchmark \citep{Pellegrini_2009_ICCV,lerner2007crowds}, utilizing the standard leave-one-out protocol where heuristics are evolved on four datasets and tested on the remaining one \citep{alahi2016social}. For all datasets, we observe 8 past frames (3.2s) to predict 12 future frames (4.8s). 



\paragraph{Baselines}
We compare \our{}-generated heuristics against heuristics and deep learning baselines. \textit{Heuristic baselines:}  suitable for resource-constrained systems, include kinematic approaches like the Constant Velocity Model (CVM), its sampling variant (CVM-S, $K=20$) \citep{scholler2020constant}, Constant Acceleration (ConstantAcc) \citep{polychronopoulos2007sensor}, Constant Turn Rate and Velocity (CTRV) \citep{lu2021ctrv}, CSCRCTR \citep{s140305239}, plus Linear Regression (LinReg) \citep{bishop2006pattern} and the physics-inspired Social Force model \citep{helbing1995social}. To benchmark against complex data-driven techniques, we include \textit{Deep learning baselines}, encompassing seminal models like Social-LSTM \citep{alahi2016social} and Social-GAN \citep{gupta2018social}, graph-based methods such as STGAT and Social-STGCNN \citep{mohamed2020social}, and more recent state-of-the-art approaches including Trajectron++ \citep{salzmann2020trajectron++}, MemoNet \citep{xu2022remember}, EigenTrajectory \citep{bae2023eigentrajectory}, and MoFlow \citep{fu2025moflow}.

\paragraph{Hardware and Software}
All experiments were conducted on a workstation equipped with an AMD Ryzen 9 7950X 16-Core Processor and a single NVIDIA RTX 3090 GPU. The \our{} framework generates trajectory prediction heuristics as executable Python code snippets in a Python 3.12 environment, employing Google's Gemini 2.0 Flash model \citep{GoogleAIDev_GeminiPricing_2025}.

\subsection{Main Results}

\begin{figure}[t!]  
    \centering
\includegraphics[width=0.72\textwidth]{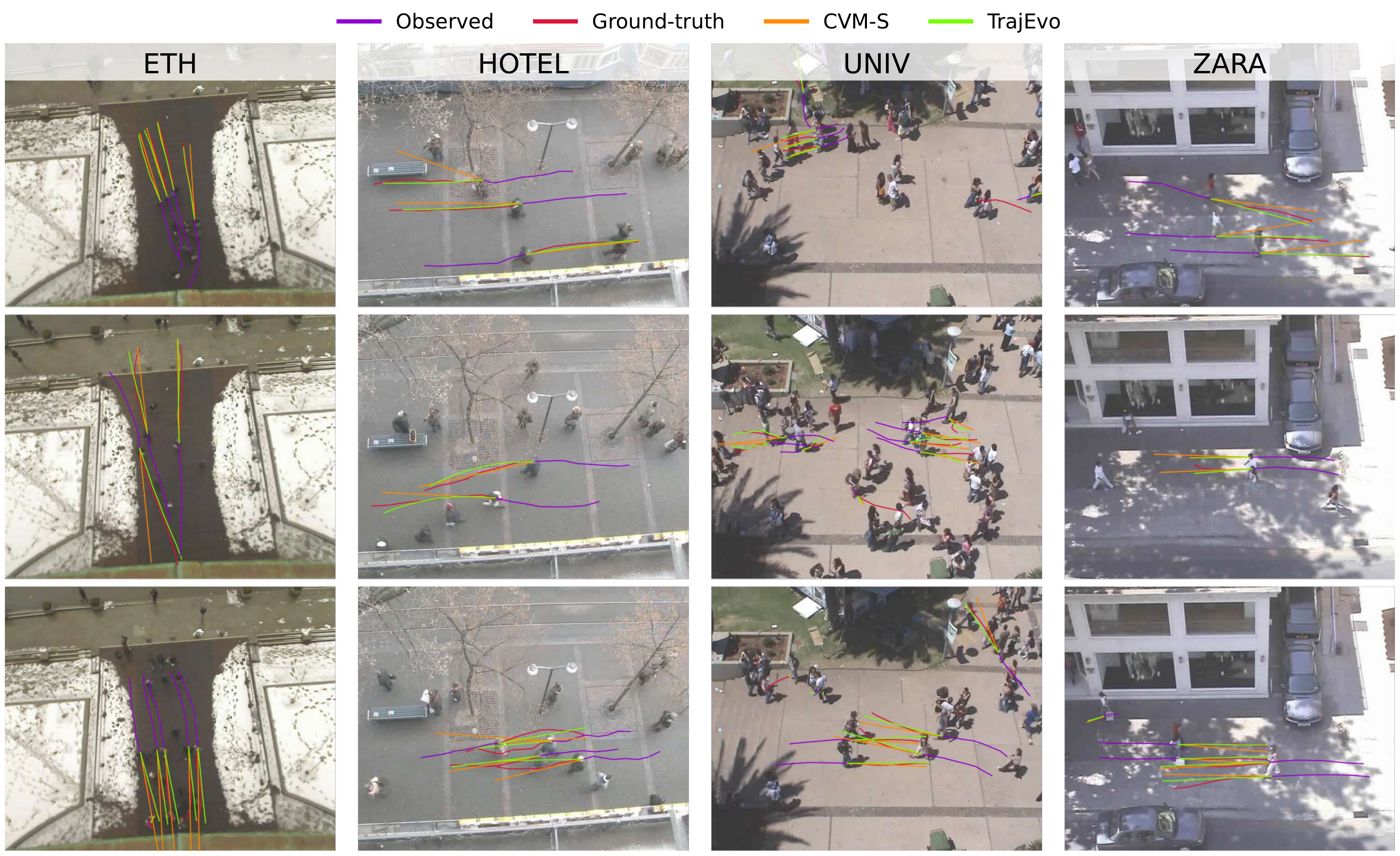}
    \caption{Comparison of trajectory prediction results between CVM-S \cite{scholler2020constant} and \our{} across different datasets. Each row illustrates a distinct behavioral pattern: (top) linear trajectories, (middle) non-linear trajectories, and (bottom) collision-avoidance cases. For each method, we report the single best trajectory out of $K=20$ samples based on the objective value $J$.}
    \label{fig:generated-trajectories-image}
    \vspace{-5mm}
\end{figure}

\paragraph{Comparison with heuristic methods}
We report results against heuristic baselines in \cref{table:main_hs}. \our{} consistently achieves the best performance across all individual ETH-UCY datasets and significantly outperforms all competitors on average. This establishes \our{} as the new state-of-the-art among heuristic approaches on this benchmark.  Interestingly, the general performance trend across baseline methods suggests that heuristics incorporating more complexity beyond basic constant velocity assumptions are generally worse suited for real-world pedestrian data, including SocialForce, with the second-best result obtained by the relatively simple CVM-S. \cref{fig:generated-trajectories-image} shows examples of generated trajectories. 
%
\input{table/main_table_hs}
%

\input{table/main_table_nn}

\paragraph{Comparison with deep learning methods} 

\cref{table:main_nn} reports results compared to deep learning methods. 
While current state-of-the-art neural networks, such as MoFlow \citep{fu2025moflow}, achieve lower average errors on the ETH-UCY benchmark, the heuristics generated by \our{} show strong performance for a non-neural approach. Notably, \our{} outperforms several established deep learning methods, including the seminal Social-LSTM \citep{alahi2016social}. Furthermore, on specific splits like ETH, \our{} surpasses even more recent complex models like Trajectron++ \citep{salzmann2020trajectron++}.

\subsection{Cross-dataset Generalization}
A crucial capability for robotic systems deployed in the real world is generalizing to situations and environments unseen during development. We evaluate this by testing models trained on ETH-UCY datasets directly on the unseen Stanford Drone Dataset (SDD) \citep{robiczek2016learning}. We report these cross-dataset generalization results in \cref{tab:cross-dataset-generalization} for three selected heuristics and three recent established neural methods. Remarkably, \our{} demonstrates superior generalization, significantly outperforming not only all heuristic baselines but also all tested deep learning methods, including SOTA models like MoFlow \citep{fu2025moflow}. \our{} performs substantially better than the best deep learning competitor, EigenTrajectory \citep{bae2023eigentrajectory}  and MoFlow. This suggests that the interpretable and efficient heuristics discovered by \our{} may possess greater robustness to domain shifts compared to complex neural networks trained on specific distributions.


\input{table/cross_dataset_generalization}

\subsection{Ablation Study} 
We conducted an ablation study, reported in \cref{tab:ablation}, to validate the effectiveness of the core components introduced in \our{}. The results demonstrate that both the Statistics Feedback Loop and Cross-Generation Elite Sampling (CGES) meaningfully improve performance. Removing either component from the full framework leads to a noticeable degradation in prediction accuracy, confirming their positive contribution. The significant impact of CGES on enhancing the quality of generated heuristics during evolution is further visualized in \cref{fig:cross-generation-elite-sampling} (right). 


\input{table/ablation}

\subsection{Analysis and Discussion}



\paragraph{Evolution Resources}
A single \our{} run takes approximately 5 minutes using Google's Gemini 2.0 Flash \citep{GoogleAIDev_GeminiPricing_2025} with an average API cost of \$0.05. In contrast, training neural methods can take a full day on a GPU, incurring significantly higher estimated compute costs -- e.g., on an RTX 3090 as reported by \citet{bae2023eigentrajectory}, around \$4 based on typical rental rates, 80$\times$ more than \our{}.


\paragraph{Inference Resources} 
\our{}-generated heuristics demonstrate significant speed advantages, requiring only 0.65ms per instance on a single CPU core. In contrast, neural baselines need 12-29ms on GPU and 248-375ms on (multi-core) CPU. Notably, \our{} achieves over a 300$\times$ speedup compared to MoFlow on CPU, highlighting its relevance for resource-constrained, real-time robotic systems. We note \our{}'s Python codes could be further compiled to optimized C++, which we leave as future work.



\paragraph{Explainability} 
Unlike neural networks, \our{} generates explainable code. For instance, the best Zara1 heuristic (see Supplementary Material) uses four sub-heuristics for its $K=20$ samples. The dominant strategy involves adaptive velocity averaging with speed-adjusted noise/variations; others use velocity rotation, memory-based avoidance, or damped extrapolation. \our{} automatically discovers such interpretable combinations of adaptive kinematics and interaction rules. The resulting heuristics' simplicity and interpretability are advantageous for robotics and autonomous driving.




\section{Conclusion}
\label{sec:conclusion}
We introduced \our{}, a novel framework leveraging Large Language Models and evolutionary algorithms to automate the design of trajectory prediction heuristics. Our experiments demonstrate that \our{} generates heuristics which not only outperform traditional methods on standard benchmarks but also exhibit superior cross-dataset generalization, remarkably surpassing even deep learning models on unseen data while remaining fast and interpretable. \our{} represents a significant first step towards automatically discovering efficient, explainable, and generalizable trajectory prediction heuristics, offering a compelling alternative in bridging the gap between handcrafted rules and complex neural networks for real-world robotic applications.

\clearpage

\section*{Limitations}





While \our{} introduces a promising paradigm for heuristic design in trajectory prediction, achieving a compelling balance of performance, efficiency, interpretability, and notably strong generalization (\cref{tab:cross-dataset-generalization}), we identify several limitations that also serve as important directions for future research:

\paragraph{In-Distribution Accuracy} 
Although \our{} significantly advances the state-of-the-art for heuristic methods and surpasses several deep learning baselines, the generated heuristics do not consistently achieve the absolute lowest error metrics on standard benchmarks compared to the most recent, highly specialized deep learning models when evaluated strictly \textit{in-distribution}. This likely reflects the inherent complexity trade-off; heuristics evolved for interpretability and speed may have a different expressivity limit compared to large neural networks. 
Future work could investigate techniques to further close this gap, potentially through more advanced evolutionary operators and heuristics integration into parts of simulation frameworks while preserving the core benefits.

\paragraph{Input Data Complexity} 
Our current evaluations focus on standard trajectory datasets using primarily positional history. Real-world robotic systems often have access to richer sensor data from which several features can be extracted, including agent types (pedestrians, vehicles), semantic maps (lanes, intersections), and perception outputs (detected obstacles, drivable space) -- for instance, given obstacle positions, we would expect \our{} to discover more likely trajectories that tend to avoid obstacles. \our{} currently does not leverage this complexity. Extending the framework to incorporate and reason about such inputs would represent a significant next step. This could enable the automatic discovery of heuristics that are more deeply context-aware and reactive to complex environmental factors.

\paragraph{Downstream Task Performance} 
We evaluate \our{} based on standard trajectory prediction metrics (minADE/minFDE). While these metrics often correlate to downstream task performance \citep{phong2023truly}, these may not always perfectly correlate with performance on downstream robotic tasks like navigation or planning. Further developing our framework to optimize heuristics directly for task-specific objectives within a closed loop (e.g., minimizing collisions or travel time in simulation) represents an interesting avenue for future works that could lead to more practically effective trajectory prediction.




\bibliography{example}  

\clearpage

\appendix

\section*{Supplementary Materials}

\section{Experimental Details}

\subsection{Hyperparameters}


The \our{} framework employs several hyperparameters that govern the evolutionary search process and the interaction with Large Language Models. Key parameters used in our experiments are detailed in \cref{tab:reevo-params}.

These include settings for population management within the evolutionary algorithm, the genetic operators, LLM-driven heuristic generation and reflection mechanisms, as well as task-specific constants for trajectory prediction evaluation. Many of these settings were determined based on common practices in evolutionary computation, adaptations from the ReEvo framework \citep{ye2024reevo}, or empirically tuned for the trajectory prediction task.

\input{table/hyperparameters}

\subsection{Detailed Resources}

\paragraph{Evolution Resources}

A single run of \our{} takes approximately 5 minutes. We employ Google's Gemini 2.0 Flash \citep{GoogleAIDev_GeminiPricing_2025}, which provides fast responses at an affordable cost. The API call costs around \$0.10 per 1M input tokens and \$0.40 per 1M output tokens, and the cost for a full run (with an average of 185,368 input tokens and 67,853 output tokens) is just around \$0.05. By comparison, recent SOTA neural methods take many more resources: for instance, \citet{bae2023eigentrajectory} reports 1 day training time on a single RTX 3090, which, as a conservative estimate for rental with market rates at the time of writing, is around 4\$.

\paragraph{Inference Resources} 
To evaluate real-time applicability for robotics, we measured inference times on SDD instances (mean over 20 runs) using both GPU and CPU. \our{}-generated heuristics are significantly faster, requiring only 0.65ms per instance on a single CPU core. In comparison, neural baselines Trajectron++, EigenTrajectory, and MoFlow need 29/19/12ms on GPU and 375/277/248ms on CPU, respectively, with their CPU inference typically utilizing all available cores\footnote{The inference time for \our{}'s Python heuristics could likely be further reduced via -- possibly \our{}-evolved -- compilation to optimized C++, which we leave as future work. As a preliminary experiment, we tried asking Claude AI to convert the generated heuristic code to C++ code, which yielded more than a 20$\times$ speedup in a zero-shot manner while resulting in the same output.}. Notably, \our{} achieves over a 300$\times$ speedup compared to MoFlow on CPU, highlighting its practical utility for resource-constrained, real-time robotic systems.

\section{Prompts}
\label{appendix:prompts}

\subsection{Common prompts}
The prompt formats are given below for the main evolutionary framework of \our{}. These are based on ReEvo \citep{ye2024reevo}, but with modified prompts tailored specifically for the task of trajectory prediction heuristic design and to incorporate the unique mechanisms of \our{}. 

\renewcommand{\lstlistingname}{Prompt}
\crefname{lstlisting}{Prompt}{Prompts}

\begin{lstlisting}[caption={System prompt for generator LLM.},  label={lst: system prompt for generator LLM}, style=promptstyle]
You are an expert in the domain of prediction heuristics. Your task is to design heuristics that can effectively solve a prediction problem.
Your response outputs Python code and nothing else. Format your code as a Python code string: "```python ... ```".    
\end{lstlisting}

\begin{lstlisting}[caption={System prompt for reflector LLM.},  label={lst: system prompt for reflector LLM}, style=promptstyle]
You are an expert in the domain of prediction heuristics. Your task is to give hints to design better heuristics.
\end{lstlisting}

\begin{lstlisting}[caption={Task description.},  label={lst: task description}, style=promptstyle]
Write a {function_name} function for {problem_description}
{function_description}
\end{lstlisting}

\begin{lstlisting}[caption={User prompt for population initialization.},  label={lst: user prompt for population initialization}, style=promptstyle]
{task_description}

{seed_function}

Refer to the format of a trivial design above. Be very creative and give `{func_name}_v2`. Output code only and enclose your code with Python code block: ```python ... ```.

{initial_long-term_reflection}
\end{lstlisting}

\begin{lstlisting}[caption={User prompt for crossover.},  label={lst: user prompt for crossover}, style=promptstyle]
{task_description}

[Worse code]
{function_signature0}
{worse_code}

[Better code]
{function_signature1}
{better_code}

[Reflection]
{short_term_reflection}

[Improved code]
Please write an improved function `{function_name}_v2`, according to the reflection. Output code only and enclose your code with Python code block: ```python ... ```.
\end{lstlisting}
\paragraph{Integration of Statistics Feedback Loop}
The prompts for reflection and mutation are designed to leverage \our{}'s Statistics Feedback Loop. \\
Specifically, the short-term reflection prompt (\cref{lst:user-prompt-short-termreflection}) and the elitist mutation prompt (\cref{lst: user prompt for elitist mutation}) explicitly require the LLM to consider "trajectory statistics" or "Code Results Analysis" when generating reflections or new heuristic code. This allows the LLM to make data-driven decisions based on the empirical performance of different heuristic strategies.

\begin{lstlisting}[caption={User prompt for short-term reflection.},  label={lst:user-prompt-short-termreflection}, style=promptstyle]
Below are two {func_name} functions for {problem_desc}
{func_desc}

You are provided with two code versions below, where the second version performs better than the first one.

[Worse code]
{worse_code}

[Worse code results analysis]
{stats_info_worse}

[Better code]
{better_code}

[Better code results analysis]

{stats_info_better}

Respond with some hints for designing better heuristics, based on the two code versions and the trajectory statistics. Be concise. Use a maximum of 200 words.
\end{lstlisting}

\begin{lstlisting}[caption={User prompt for long-term reflection.},  label={lst: user prompt for long-term reflection}, style=promptstyle]
Below are two {function_name} functions for {problem_description}
{function_description}

You are provided with two code versions below, where the second version performs better than the first one.

[Worse code]
{worse_code}

[Better code]
{better_code}

You respond with some hints for designing better heuristics, based on the two code versions and using less than 20 words.
\end{lstlisting}

\begin{lstlisting}[caption={User prompt for elitist mutation.},  label={lst: user prompt for elitist mutation}, style=promptstyle]
{user_generator}

[Prior reflection]
{reflection}

[Code]
{func_signature1}
{elitist_code}

[Code Results Analysis]
{stats_info_elitist}

[Improved code]
Please write a mutated function `{func_name}_v2`, according to the reflection. Output code only and enclose your code with Python code block: ```python ... ```. 

Please generate mutation versions that are significantly different from the base code to increase exploration diversity.
\end{lstlisting}

\subsection{Trajectory Prediction-specific Prompts}

\paragraph{Domain Specialization} All prompts are contextualized for the domain of trajectory prediction heuristics. For example, the system prompts are deeply contextualized through specific prompts detailing the trajectory prediction problem:

\begin{lstlisting}[caption={Function Signature},  label={lst: function signature}, style=promptstyle]]
def predict_trajectory{version}(trajectory: np.ndarray) -> np.ndarray:
\end{lstlisting}

\begin{lstlisting}[caption={Function Description},  label={lst: function description}, style=promptstyle]]
The predict_trajectory function takes as input the current trajectory (8 frames) and generates 20 possible future trajectories for the next 12 frames. It has only one parameter: the past trajectory array.
The output is a numpy array of shape [20, num_agents, 12, 2] containing all 20 trajectories.
Note that we are interesting in obtaining at least one good trajectory, not necessarily 20.
Thus, diversifying a little bit is good.
Note that the heuristic should be generalizable to new distributions.
\end{lstlisting}

\begin{lstlisting}[caption={Seed Function}, label={lst: seed function}, style=promptstyle]]
def predict_trajectory(trajectory):
    """Generate 20 possible future trajectories
    Args:
        - trajectory [num_agents, traj_length, 2]: here the traj_length is 8;
    Returns:
        - 20 diverse trajectories [20, num_agents, 12, 2]
    """
    all_trajectories = []
    for _ in range(20):
        current_pos = trajectory[:, -1, :]
        velocity = trajectory[:, -1, :] - trajectory[:, -2, :] # only use the last two frames
        predictions = []
        for t in range(1, 12+1): # 12 future frames
            current_pos = current_pos + velocity * 1 # dt
            predictions.append(current_pos.copy())
        pred_trajectory = np.stack(predictions, axis=1)
        all_trajectories.append(pred_trajectory)
    all_trajectories = np.stack(all_trajectories, axis=0)
    return all_trajectories
\end{lstlisting}

\begin{lstlisting}[caption={External Knowledge},  label={lst: external knowledge}, style=promptstyle]]
# External Knowledge for Pedestrian Trajectory Prediction

## Task Definition
- We are using the ETH/UCY dataset for this task (human trajectory prediction)
- Input: Past 8 frames of pedestrian positions
- Output: Future 12 frames of pedestrian positions
- Variable number of pedestrians per scene
\end{lstlisting}

\section{\our{} Output}

\subsection{Generated Heuristics}

\renewcommand{\lstlistingname}{Heuristic}
\crefname{lstlisting}{Heuristic}{Heuristics}

\input{llm/generated_heuristic}

\subsection{Reflections}

\renewcommand{\lstlistingname}{Output}
\crefname{lstlisting}{Output}{Outputs}

\begin{lstlisting}[caption={Long-term reasoning output},  label={lst:reasoning}, style=promptstyle]]
Based on comparative analysis, prioritize these heuristics:

1.  **Hierarchical Stochasticity:** Sample trajectory-level parameters (speed scale, movement pattern) *once* per trajectory. Then, apply agent-specific stochastic variations within those constraints. Introduce `global_randomness` sampled *once* per trajectory to couple different parameters.
2.  **Adaptive Movement Primitives:** Condition movement model probabilities (stop, turn, straight, lane change, obstacle avoidance) on agent state (speed, acceleration, past turning behavior, context). Consider longer history windows.
3.  **Refine Noise & Parameters:** Finetune noise scales and apply dampening. Experiment with learnable parameters and wider ranges. Directly manipulate velocity and acceleration stochastically for smoother transitions.
4.  **Contextual Interactions:** Enhance social force models, considering intentions, agent types, and environment.
5.  **Guaranteed Diversity:** Ensure movement probabilities sum to 1.
6.  **Post Processing:** Apply smoothing and collision avoidance.
7. **Intentions:** Incorporate high level intentions such as "going to an area."
\end{lstlisting}

\cref{lst:reasoning}  shows an example of long-term reasoning output for the model, based on the comparative analysis. 
\our{} discovers several interesting heuristics for trajectory forecasting, such as applying diverse noise factors, social force models, diversity, and modeling intentions to model possible future trajectories.

\cref{lst: interactions} shows some more outputs of \our{} from various runs, which discovers some interesting helper functions that model interactions such as stochastic, social force, and diversity.

\begin{lstlisting}[caption={Selected \our{} Interactions}, label={lst: interactions}, language=Python, style=heuristicstyle]]
################### 
# Model with Noise
################### 

def acceleration_model_with_noise(trajectory, noise_level_base=0.05, prediction_steps=12):
    velocity = trajectory[:, -1, :] - trajectory[:, -2, :]
    acceleration = velocity - (trajectory[:, -2, :] - trajectory[:, -3, :])
    current_pos = trajectory[:, -1, :].copy()
    current_velocity = velocity.copy()
    predictions = []
    for i in range(prediction_steps):
        noise_level = noise_level_base * (i + 1)
        current_velocity = current_velocity + acceleration + np.random.normal(0, noise_level, size=current_velocity.shape)
        current_pos = current_pos + current_velocity
        predictions.append(current_pos.copy())
    return np.stack(predictions, axis=1)

################### 
# Social Force
################### 

def constant_velocity_with_repulsion(trajectory, get_nearby_agents, repulsion_strength=0.05, num_steps=12):
    velocity = trajectory[:, -1, :] - trajectory[:, -2, :]
    current_pos = trajectory[:, -1, :].copy()
    repulsion_force = np.zeros_like(current_pos)
    num_agents = trajectory.shape[0]

    for _ in range(num_steps):
        for agent_index in range(num_agents):
            nearby_agents = get_nearby_agents(agent_index, current_pos)
            for neighbor_index in nearby_agents:
                direction = current_pos[agent_index] - current_pos[neighbor_index]
                distance = np.linalg.norm(direction)
                if distance > 0:
                    repulsion_force[agent_index] += (direction / (distance**2 + 0.001)) * repulsion_strength
    return repulsion_force

################### 
# Diversity
################### 

def simple_random_walk(trajectory, noise_level_base=0.2, prediction_steps=12):
    current_pos = trajectory[:, -1, :].copy()
    predictions = []
    for _ in range(prediction_steps):
        noise_level = noise_level_base * (_ + 1)
        current_pos = current_pos + np.random.normal(0, noise_level, size=current_pos.shape)
        predictions.append(current_pos.copy())
    return np.stack(predictions, axis=1)
\end{lstlisting}

\end{document}

%% file: figures/cross-generation-elite-sampling.tex
\begin{figure}[h!] 
    \centering 
    \begin{subfigure}[b]{0.33\textwidth} 
        \centering
        \includegraphics[width=\linewidth]{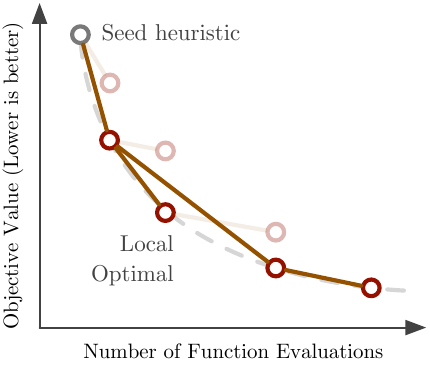}
    \end{subfigure}
    %
    \hspace{10mm}
    \begin{subfigure}[b]{0.39\textwidth} 
        \centering
        \includegraphics[width=\linewidth]{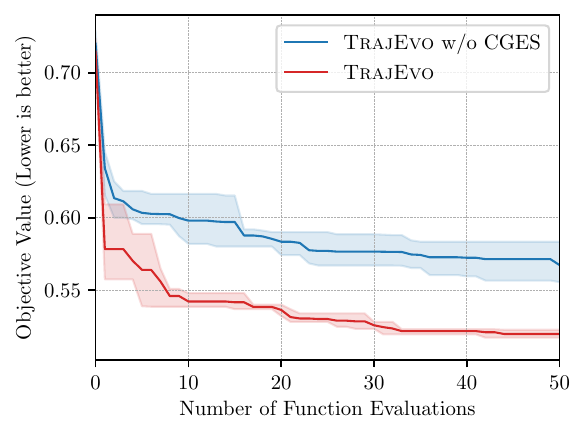}
    \end{subfigure}
\vspace{-2mm}
    \caption{Cross-Generation Elite Sampling (CGES) helps escape local optima by sampling elite individuals from past generations (left), which greatly helps achieve much better objective values (right).}
    \label{fig:cross-generation-elite-sampling}
\end{figure}%

%% file: figures/trajectory-analysis.tex
\begin{figure}[h!]
\begin{minipage}[t]{0.48\textwidth}
\begin{lstlisting}[label={lst:stats}, style=promptstyle]
...
<stats>
Statistics of trajectory index counts with the lowest ADE. These help us understand which heuristics contribute to the performance for at least some trajectories. 
Traj Index: Count: {0: 67, 1: 10, 2: 3, 3: 2, 4: 7, 5: 0, ... 18: 0, 19: 2}
</stats>
...
\end{lstlisting}
\end{minipage}
\hfill
\begin{minipage}[t]{0.48\textwidth}
\begin{lstlisting}[label={lst:analysis}, style=promptstyle]
...
The better code explicitly includes a deterministic (linear extrapolation) trajectory as the first prediction. This significantly improves ADE, as evidenced by trajectory 0 having the lowest ADE count by far. The statistics suggest it's beneficial to have at least one trajectory that closely follows the established motion.
...
\end{lstlisting}
\end{minipage}
\vspace{-2mm}
\caption{Statistics obtained by running \our{}-generated code are provided alongside the corresponding code as an input to the reflector (left). \our{} then analyses this and gathers insights into how to evolve better trajectory prediction heuristics (right).}
\label{fig:trajectory-analysis}
\end{figure}

%% file: table/main_table_hs.tex
\begin{table*}[t]
\centering
\caption{Comparison of \our{} with trajectory prediction heuristics across datasets with mean minADE$_{20}$ / minFDE$_{20}$ (meters) on the ETH-UCY dataset.}
\vspace{-2mm}
\begin{tabular}{l||c|c|c|c|c||c}
\toprule
Method & ETH & HOTEL & UNIV & ZARA1 & ZARA2 & AVG \\
\midrule
CVM \cite{scholler2020constant}   & 1.01/2.24 & 0.32/0.61 & 0.54/1.21 & 0.42/0.95 & 0.33/0.75 & 0.52/1.15 \\
CVM-S \cite{scholler2020constant}  & 0.92/2.01 & 0.27/0.51 & 0.53/1.17 & 0.37/0.77 & 0.28/0.63 & 0.47/1.02 \\
ConstantAcc \citep{polychronopoulos2007sensor}  & 3.12/7.98 & 1.64/4.19 & 1.02/2.60 & 0.81/2.05 & 0.60/1.53 & 1.44/3.67 \\
CTRV \cite{lu2021ctrv} & 1.62/3.64 & 0.72/1.09 & 0.71/1.59 & 0.65/1.50 & 0.48/1.10 & 0.84/1.78 \\
CSCRCTR \cite{s140305239} & 2.27/4.61 & 1.03/2.18 & 1.35/3.12& 0.96/2.12 & 0.90/2.10  & 1.30/2.83 \\
LinReg \citep{bishop2006pattern} &  1.04/2.20 & 0.26/0.47 & 0.76/1.48 & 0.62/1.22 & 0.47/0.93 & 0.63/1.26 \\
SocialForce \cite{helbing1995social} & 1.46/2.48 & 0.69/1.23 & 0.96/1.75 & 1.37/2.51 & 0.84/1.53 & 1.06/1.90 \\
\midrule
\our{}  
& \textbf{0.47/0.78} 
& \textbf{0.17/0.31} 
& \textbf{0.52/1.10} 
& \textbf{0.36/0.77} 
& \textbf{0.28/0.58} 
& \textbf{0.36/0.71} \\
\bottomrule
\end{tabular}
\vspace{-2mm}
\label{table:main_hs}
\end{table*}

%% file: table/main_table_nn.tex
\begin{table*}[t]
\centering
\caption{Comparison of \our{} with deep learning approaches (mean minADE$_{20}$/minFDE$_{20}$ on ETH-UCY). Each \hl{highlighted} number is one where \our{}-generated heuristic outperforms that model.}
\vspace{-2mm}
\resizebox{\textwidth}{!}{%
\begin{tabular}{l||c|c|c|c|c||c}
\toprule
Method & ETH & HOTEL & UNIV & ZARA1 & ZARA2 & AVG \\
\midrule
Social-LSTM \cite{alahi2016social}
  & \hl{1.09}/\hl{2.35}
  & \hl{0.79}/\hl{1.76}
  & \hl{0.67}/\hl{1.40}
  & \hl{0.56}/\hl{1.17}
  & \hl{0.72}/\hl{1.54}
  & \hl{0.77}/\hl{1.64} \\

Social-GAN \cite{gupta2018social}
  & \hl{0.87}/\hl{1.62}
  & \hl{0.67}/\hl{1.37}
  & \hl{0.76}/\hl{1.52}
  & 0.35/0.68
  & \hl{0.42}/\hl{0.84}
  & \hl{0.61}/\hl{1.21} \\

STGAT \cite{Huang_2019_ICCV}
  & \hl{0.65}/\hl{1.12}
  & \hl{0.35}/\hl{0.66}
  & \hl{0.52}/1.10
  & 0.34/0.69
  & \hl{0.29}/\hl{0.60}
  & \hl{0.43}/\hl{0.83} \\

Social-STGCNN \cite{Huang_2019_ICCV}
  & \hl{0.64}/\hl{1.11}
  & \hl{0.49}/\hl{0.85}
  & 0.44/0.79
  & 0.34/0.53
  & \hl{0.30}/0.48
  & \hl{0.44}/\hl{0.75} \\

Trajectron++ \cite{salzmann2020trajectron++}
  & \hl{0.61}/\hl{1.03}
  & \hl{0.20}/0.28
  & 0.30/0.55
  & 0.24/0.41
  & 0.18/0.32
  & 0.31/0.52 \\

MemoNet \cite{xu2022remember}
  & 0.41/0.61
  & \textbf{0.11}/\textbf{0.17}
  & 0.24/0.43
  & 0.18/0.32
  & 0.14/0.24
  & 0.21/0.35 \\



EigenTrajectory \cite{bae2023eigentrajectory}
  & \textbf{0.36}/\textbf{0.53}
  & 0.12/0.19
  & 0.24/0.43
  & 0.19/0.33
  & 0.14/0.24
  & 0.21/0.34 \\


MoFlow \cite{fu2025moflow}
  & 0.40/0.57
  & \textbf{0.11}/\textbf{0.17}
  & \textbf{0.23}/\textbf{0.39}
  & \textbf{0.15}/\textbf{0.26}
  & \textbf{0.12}/\textbf{0.22}
  & \textbf{0.20}/\textbf{0.32} \\

\midrule
\our{}
& {0.47/0.78} 
& {0.17/0.31} 
& {0.52/1.10} 
& {0.36/0.77} 
& {0.28/0.58} 
& {0.36/0.71} \\
\bottomrule
\end{tabular}%
}
\label{table:main_nn}
\end{table*}

%% file: table/cross_dataset_generalization.tex
\begin{table*}[t]
\centering
\caption{Cross-dataset generalization of methods trained on different ETH-UCY splits and tested on the unseen SDD dataset. We report minADE$_{20}$ / minFDE$_{20}$ (pixels) on the SDD dataset.}
\vspace{-2mm}
\label{tab:cross-dataset-generalization}
\resizebox{\textwidth}{!}{%
\begin{tabular}{l||c|c|c|c|c||c}
\toprule
\multirow{2}{*}{Method} & ETH & HOTEL & UNIV & ZARA1 & ZARA2 & \multirow{2}{*}{AVG} \\
 & $\rightarrow$ SDD & $\rightarrow$ SDD & $\rightarrow$ SDD & $\rightarrow$ SDD & $\rightarrow$ SDD & \\
\midrule
SocialForce \cite{helbing1995social} 
& 33.64/60.63
& 33.64/60.63
& 33.64/60.63 
& 33.64/60.63 
& 33.64/60.63 
& 33.64/60.63 \\
CVM \cite{scholler2020constant}
  & 18.82/37.95
  & 18.82/37.95
  & 18.82/37.95
  & 18.82/37.95
  & 18.82/37.95
  & 18.82/37.95\\
CVM-S \cite{scholler2020constant}
  & 16.28/31.84
  & 16.28/31.84
  & 16.28/31.84
  & 16.28/31.84
  & 16.28/31.84
  & 16.28/31.84 \\
  \midrule
Trajectron++ \cite{salzmann2020trajectron++}
  & 46.72/69.11
  & 47.30/67.76
  & 46.08/75.90
  & 47.30/72.19
  & 46.78/68.59
  & 46.84/70.71 \\
EigenTrajectory \cite{bae2023eigentrajectory}
  & 14.51/25.13
  & 14.69/24.64
  & 14.31/27.60
  & 14.69/26.25
  & 14.53/24.94
  & 14.55/25.71 \\
 MoFlow \cite{fu2025moflow}
& 17.00/27.98
& 17.21/27.43
& 17.00/30.63
& 17.24/29.22
& 17.27/27.56
& 17.14/28.56 \\
  \midrule
\our{}
  & \textbf{12.61}/\textbf{23.84}
  & \textbf{12.56}/\textbf{24.02}
  & \textbf{12.68}/\textbf{23.63}
  & \textbf{13.21}/\textbf{25.65}
  & \textbf{12.18}/\textbf{23.54}
  & \textbf{12.65}/\textbf{24.14} \\
\bottomrule
\end{tabular}%
}
\vspace{-3mm}
\label{table:cross_dataset_generalization}
\end{table*}

%% file: table/ablation.tex
\begin{table}[t]
\caption{Ablation study for the evolution framework removing different components $(\downarrow)$.}
\label{tab:ablation}
\centering
\resizebox{\linewidth}{!}{
\begin{tabular}{@{}lccccc@{}}
\toprule
{Method} & {ETH} & {HOTEL} & {UNIV} & {ZARA1} & {ZARA2}\\
\midrule
\our{} (full)
& \textbf{0.47/0.78} 
& \textbf{0.17/0.31} 
& \textbf{0.52/1.10} 
& \textbf{0.36/0.77} 
& \textbf{0.28/0.58} \\
~ - Statistics Feedback Loop 
& 0.59/1.13
& 0.19/0.34 
& 0.53/1.13
& 0.37/0.77
& 0.28/0.60
\\
\quad ~ -  Cross-Generation Elite Sampling
& 0.68/1.37
& 0.26/0.46
&0.60/1.22
&0.37/0.79
& 0.32/0.66
\\
\bottomrule
\end{tabular}
}
\end{table}

%% file: table/hyperparameters.tex

\begin{table}[htbp] 
\centering
\caption{Main hyperparameters for the \our{} framework.}
\label{tab:reevo-params} 
\begin{tabular}{@{}lll@{}}
\toprule
\textbf{Category} & \textbf{Hyperparameter} & \textbf{Value} \\
\midrule
\multirow{6}{*}{Evolutionary Algorithm} & Population size  & 10 \\
& Number of initial generation & 8 \\
& Elite ratio for crossover  & 0.3 \\
& Crossover rate & 1 \\
& Mutation rate & 0.5 \\
& CGES Softmax temperature  & 1.0 \\
\midrule
\multirow{4}{*}{LLM} & LLM model & Gemini 2.0 Flash \\
& LLM temperature (generator and reflector) & 1 \\
& Max words for short-term reflection & 200 words \\
& Max words for long-term reflection & 20 words \\
\midrule
\multirow{5}{*}{Trajectory Prediction} & Objective $J$ ADE weight ($w_{\text{ADE}}$) & 0.6 \\
& Objective $J$ FDE weight ($w_{\text{FDE}}$) & 0.4 \\
& Num. prediction samples ($K$) & 20 \\
& Observation length ($T_{\text{obs}}$) & 8 frames (3.2s) \\
& Prediction length ($T_{\text{pred}}$) & 12 frames (4.8s) \\
\bottomrule
\end{tabular}%
\label{tab:reevo-params}
\end{table}

%% file: llm/generated_heuristic.tex
\begin{lstlisting}[caption={The best \our{}-generated heuristic for Zara 1.},  label={lst:heuristic-eth}, language=Python, style=heuristicstyle]
import numpy as np

def predict_trajectory(trajectory: np.ndarray) -> np.ndarray:
    """Generate 20 possible future trajectories with enhanced diversification and adaptive strategies.

    Args:
        trajectory (np.ndarray): [num_agents, traj_length, 2] where traj_length is 8.

    Returns:
        np.ndarray: 20 diverse trajectories [20, num_agents, 12, 2].
    """
    num_agents = trajectory.shape[0]
    all_trajectories = []
    history_len = trajectory.shape[1]

    for i in range(20):
        current_pos = trajectory[:, -1, :]

        # Option 1: Dominant strategy - Average velocity with adaptive noise, rotation, and parameter variation
        if i < 14:  # Increased to 14, best performing strategy
            velocity = np.zeros_like(current_pos)
            weights_sum = 0.0
            decay_rate = np.random.uniform(0.1, 0.3)  # Adaptive decay rate
            for k in range(min(history_len - 1, 5)):
                weight = np.exp(-decay_rate * k)
                velocity += weight * (trajectory[:, -1 - k, :] - trajectory[:, -2 - k, :])
                weights_sum += weight
            velocity /= (weights_sum + 1e-8)

            avg_speed = np.mean(np.linalg.norm(velocity, axis=1))
            noise_scale = 0.012 + avg_speed * 0.008
            noise = np.random.normal(0, noise_scale, size=(num_agents, 12, 2))

            angle = np.random.uniform(-0.05, 0.05)
            rotation_matrix = np.array([[np.cos(angle), -np.sin(angle)],
                                        [np.sin(angle), np.cos(angle)]])
            velocity = velocity @ rotation_matrix

            momentum = 0.0
            jerk_factor = 0.0
            damping = 0.0

            # Parameter Variation
            if i % 6 == 0:
                noise_scale *= np.random.uniform(0.9, 1.1)  # Fine-tuned noise scale variation
                noise = np.random.normal(0, noise_scale, size=(num_agents, 12, 2))
            elif i % 6 == 1:
                angle_scale = 0.06 + avg_speed * 0.02
                angle = np.random.uniform(-angle_scale * np.random.uniform(0.8, 1.2), angle_scale * np.random.uniform(0.8, 1.2))
                rotation_matrix = np.array([[np.cos(angle), -np.sin(angle)],
                                            [np.sin(angle), np.cos(angle)]])
                velocity = velocity @ rotation_matrix
            elif i % 6 == 2:
                momentum = np.random.uniform(0.06, 0.14)  # Vary momentum
                velocity = momentum * velocity + (1 - momentum) * (trajectory[:, -1, :] - trajectory[:, -2, :])
            elif i % 6 == 3:  # Add jerk
                jerk_factor = np.random.uniform(0.0025, 0.0065)
                if history_len > 2:
                    jerk = (trajectory[:, -1, :] - 2 * trajectory[:, -2, :] + trajectory[:, -3, :])
                else:
                    jerk = np.zeros_like(velocity)
                velocity += jerk_factor * jerk
            elif i % 6 == 4: # Damping
                damping = np.random.uniform(0.006, 0.019)
                velocity = velocity * (1 - damping)
            else: # Adaptive Noise Scale
                noise_scale = 0.01 + avg_speed * np.random.uniform(0.006, 0.014)
                noise = np.random.normal(0, noise_scale, size=(num_agents, 12, 2))

            predictions = []
            for t in range(1, 13):
                current_pos = current_pos + velocity + noise[:, t-1, :] / (t**0.4)
                predictions.append(current_pos.copy())
            pred_trajectory = np.stack(predictions, axis=1)

        # Option 2: Velocity rotation with adaptive angle
        elif i < 17:  # Increased to 17.
            velocity = trajectory[:, -1, :] - trajectory[:, -2, :]
            avg_speed = np.mean(np.linalg.norm(velocity, axis=1))
            angle_scale = 0.13 + avg_speed * 0.05  # adaptive angle

            angle = np.random.uniform(-angle_scale, angle_scale)  # adaptive range
            rotation_matrix = np.array([[np.cos(angle), -np.sin(angle)],
                                        [np.sin(angle), np.cos(angle)]])
            velocity = velocity @ rotation_matrix

            noise_scale = 0.007 + avg_speed * 0.004
            noise = np.random.normal(0, noise_scale, size=(num_agents, 12, 2))

            predictions = []
            for t in range(1, 13):
                current_pos = current_pos + velocity + noise[:, t-1, :] / (t**0.5)
                predictions.append(current_pos.copy())
            pred_trajectory = np.stack(predictions, axis=1)

        # Option 3: Memory-based approach (repeating last velocity) + Enhanced Collision Avoidance
        elif i < 19:  # Increased to 19
            velocity = trajectory[:, -1, :] - trajectory[:, -2, :]
            # Enhanced smoothing with more velocity history
            if history_len > 3:
                velocity = 0.55 * velocity + 0.3 * (trajectory[:, -2, :] - trajectory[:, -3, :]) + 0.15 * (trajectory[:, -3, :] - trajectory[:, -4, :])
            elif history_len > 2:
                velocity = 0.65 * velocity + 0.35 * (trajectory[:, -2, :] - trajectory[:, -3, :])
            else:
                velocity = velocity # do nothing

            avg_speed = np.mean(np.linalg.norm(velocity, axis=1))

            # Adaptive Laplacian noise
            noise_scale = 0.005 + avg_speed * 0.0015
            noise = np.random.laplace(0, noise_scale, size=(num_agents, 2))
            velocity = velocity + noise

            # Enhanced collision avoidance
            repulsion_strength = 0.0011 # Adjusted repulsion strength
            predictions = []
            temp_pos = current_pos.copy()

            # Store predicted positions for efficient collision calculation at each timestep
            future_positions = [temp_pos.copy()]  # Start with current position
            for t in range(1, 13):
                net_repulsions = np.zeros_like(temp_pos)
                for agent_idx in range(num_agents):
                    for other_idx in range(num_agents):
                        if agent_idx != other_idx:
                            direction = temp_pos[agent_idx] - temp_pos[other_idx]
                            distance = np.linalg.norm(direction)
                            if distance < 1.05:  # Adjusted interaction threshold
                                repulsion = (direction / (distance + 1e-6)) * repulsion_strength * np.exp(-distance)  # distance-based decay
                                net_repulsions[agent_idx] += repulsion

                velocity = 0.9 * velocity + 0.1 * net_repulsions # Damping the change in velocity
                temp_pos = temp_pos + velocity
                future_positions.append(temp_pos.copy()) # Store for future repulsion calculations
                predictions.append(temp_pos.copy())

            pred_trajectory = np.stack(predictions, axis=1)

        # Option 4: Linear prediction with adaptive damping and larger noise.
        else:
            velocity = trajectory[:, -1, :] - trajectory[:, -2, :]
            damping = np.random.uniform(0.017, 0.038)  # damping factor

            noise_scale = 0.028
            noise = np.random.normal(0, noise_scale, size=(num_agents, 12, 2))

            predictions = []
            for t in range(1, 13):
                velocity = velocity * (1-damping) + noise[:, t-1, :] / (t**0.4)  # damping
                current_pos = current_pos + velocity
                predictions.append(current_pos.copy())
            pred_trajectory = np.stack(predictions, axis=1)

        all_trajectories.append(pred_trajectory)

    all_trajectories = np.stack(all_trajectories, axis=0)
    return all_trajectories
\end{lstlisting}